\documentclass[11pt,a4paper]{article}
\usepackage[hyperref]{emnlp-ijcnlp-2019}
\usepackage{times}
\usepackage{graphicx}
\usepackage{latexsym}
\usepackage{booktabs}
\usepackage{url}
\usepackage{appendix}

\aclfinalcopy % Uncomment this line for the final submission

\title{SimpleBooks: Long-term dependency book dataset with simplified English vocabulary for word-level language modeling}

\author{Huyen Nguyen \\
  NVIDIA \\
  {\tt chipn@nvidia.com} \\}

\date{}

\begin{document}
\maketitle
\begin{abstract}
  With language modeling becoming the popular base task for unsupervised representation learning in Natural Language Processing, it is important to come up with new architectures and techniques for faster and better training of language models (LMs). However, due to a peculiarity of languages -- the larger the dataset, the higher the average number of times a word appears in that dataset -- datasets of different sizes have very different properties. Architectures performing well on small datasets might not perform well on larger ones. For example, LSTM models perform well on WikiText-2 but poorly on WikiText-103, while Transformer models perform well on WikiText-103 but not on WikiText-2. For setups like architectural search, this is a challenge since it is prohibitively costly to run a search on the full dataset but it is not indicative to experiment on smaller ones. In this paper, we introduce SimpleBooks, a small dataset with the average word frequency as high as that of much larger ones. Created from 1,573 Gutenberg books with the highest ratio of word-level book length to vocabulary size, SimpleBooks contains 92M word-level tokens, on par with WikiText-103 (103M tokens), but has the vocabulary of 98K, a third of WikiText-103's. SimpleBooks can be downloaded from \url{https://dldata-public.s3.us-east-2.amazonaws.com/simplebooks.zip}.
\end{abstract}

\section{Introduction}
To develop new architectures or techniques in deep learning, it is important to have small and easy-to-train datasets for quick experiments. Datasets like MNIST \citep{cirecsan2012multi}, Fashion-MNIST \citep{xiao2017fashion}, and CIFAR \citep{krizhevsky2009learning} have become the standard testbeds in the field of computer vision. Their contributions are invaluable.

Given how popular the task of language modeling has become, it is important to have a small long-term dependency dataset that is representative of bigger datasets to serve as a testbed and benchmark for language modeling task. However, this is hard to achieve due to one peculiarity of languages: the larger the body of text, the higher the average number of times a word appears in that text. For simplicity, let FREQ denote the average number of times a token appears in a dataset.

Consider the most popular datasets for word-level LMs:
\begin{itemize}
    \item \textbf{Penn TreeBank} (PTB) dataset contains the Penn Treebank portion of the Wall Street Journal corpus, pre-processed by Mikolov et al. \citep{mikolov2011empirical}. It consists of 929k tokens for train, 73k for validation, and 82k for test. All words are lower-cased, numbers replaced with N, and most punctuations removed. The vocabulary is the most frequent 10k words. Out-of-vocabulary (OOV) words are replaced by an $<$unk$>$ token. PTB contains sentences instead of paragraphs, so its context is limited.
    \item \textbf{WikiText-103} consists of 28,475 good and featured articles from Wikipedia. It has long-term dependency with 103 million tokens. After replacing all tokens that appear less than 3 times with a $<$unk$>$ token, it has a vocabulary size of 267,735. \citep{merity2016pointer} This makes it prohibitive to experiment with word-level LMs on this dataset. For an embedding size of 400, the embedding layer alone has \textbf{267K x 400 $\approx$ 106M} parameters.
    \item \textbf{WikiText-2} is a 2M token version of WikiText-103 with a vocabulary size of 33,278.
    \item \textbf{One-Billion Word} (1Billion) dataset consists of 829M tokens over a vocabulary of 793K. Sentences in this dataset are shuffled and hence the context is limited. It is also too big for quick experimenting.
\end{itemize}

Table \ref{tab:all-datasets} shows that the bigger the body of text, the higher FREQ. The low FREQ for PTB and WikiText-2 explains why it is so hard to achieve low perplexity on these two datasets: each token simply does not appear enough times for the language model to learn a good representation of each token. The high percentage of OOV tokens also adds to the difficulty.

\begin{figure}[t]
  \centering
  \includegraphics[scale=0.23]{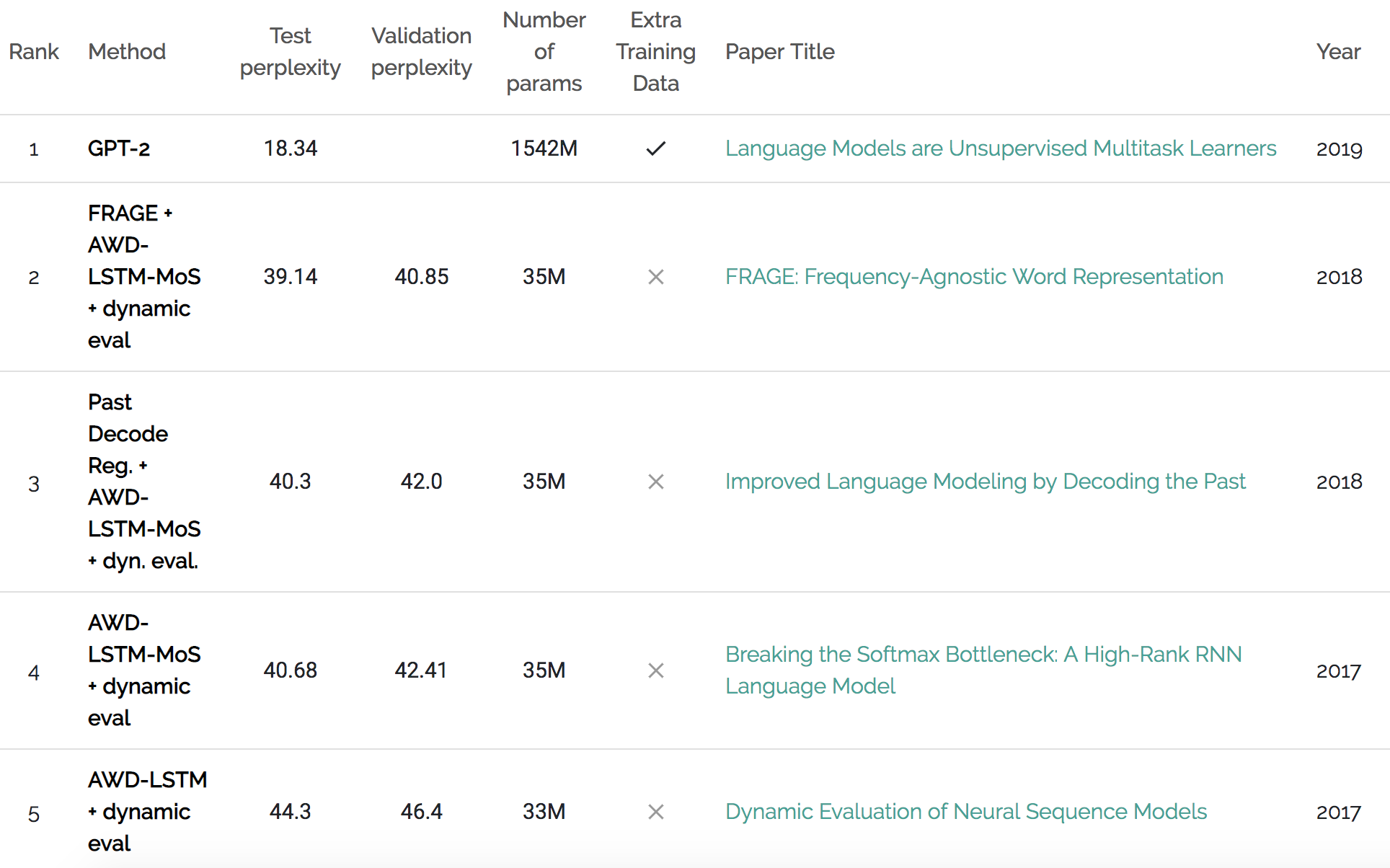}
  \vspace{-3mm}
  \caption{Top 4 performing models on WikiText-2 without external data are all LSTM-based. From \url{papersiwthcode.com}.}
  \label{fig:wkt2-sota}
\end{figure}

\begin{figure}[t]
  \centering
  \includegraphics[scale=0.23]{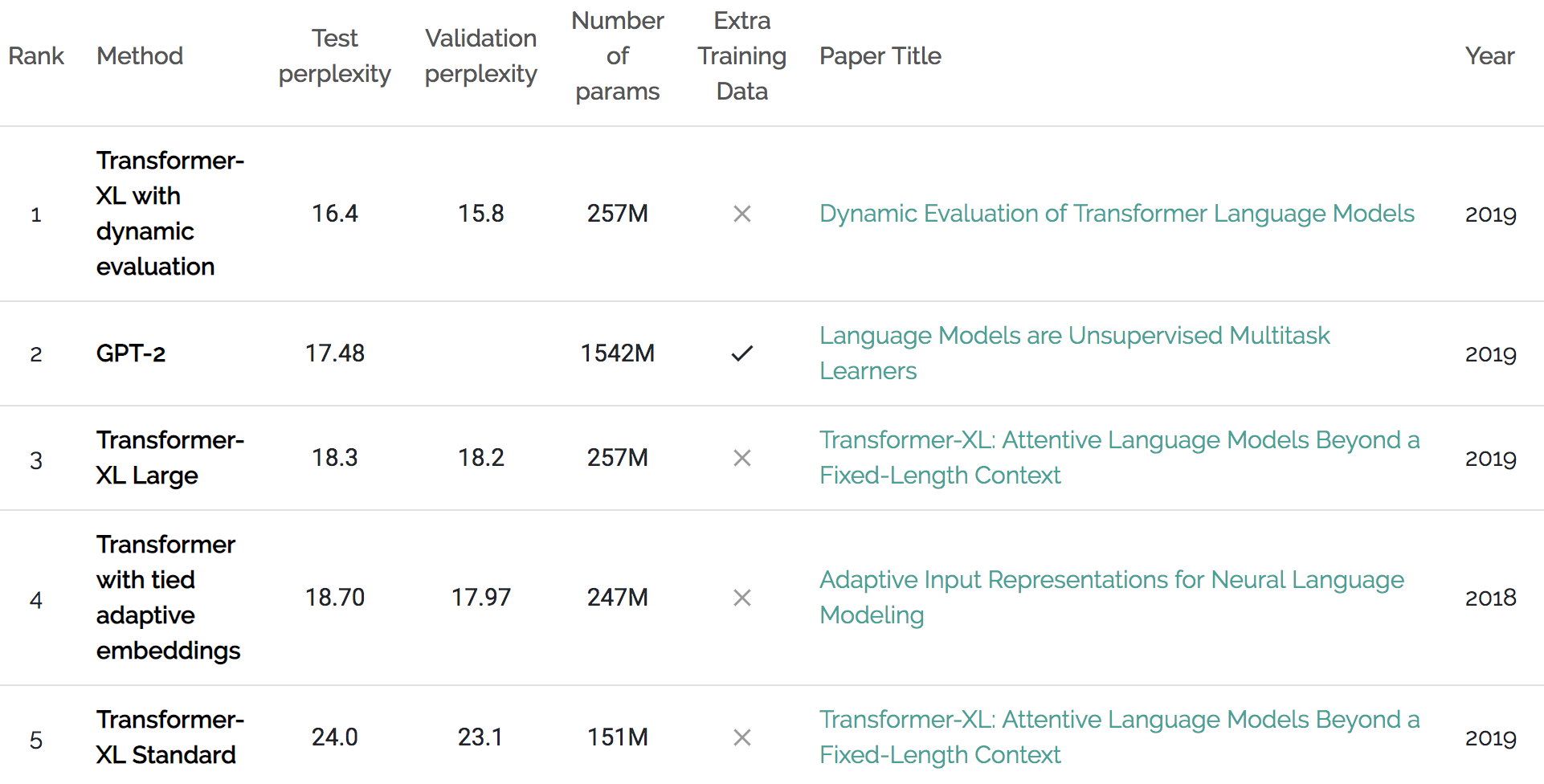}
  \vspace{-3mm}
  \caption{Top 4 performing models on WikiText-103 without external data are all Transformer-based. From \url{papersiwthcode.com}.}
  \label{fig:wkt103-sota}
\end{figure}

Looking at the state-of-the-art (SOTA) results, there is a pattern: the best performing models on small datasets like PTB and WikiText-2 are LSTM-based while the best performing models on larger datasets like WikiText-103 and 1Billion are dominated by Transformer models (See Figures \ref{fig:wkt2-sota} and \ref{fig:wkt103-sota}).

There are a few possible reasons. One is because LSTMs have been around longer, there have been more regularization techniques developed for them, which make them work better with small datasets that often require more regularization.

Another is that for datasets with low FREQ, models have to rely more on the structural information of text, and RNNs are better at capturing and exploiting hierarchical information \citep{tran2018importance}. RNNs, due to their recurrent nature, have a stronger inductive bias towards the most recent symbols. Transformer models, since they can attend to any symbol within the context, need a lot of data to learn that the most recent symbols are more relevant. When incorporating inductive bias, Transformer models seem to generalize better on small datasets \citep{dehghani2018universal}.

One thing is clear: an architecture that works well for a small dataset might not work well for a bigger one. This makes it challenging for setups like architectural search where it is prohibitive to run the search on a large dataset, yet architectures found by the search on a small dataset might not be useful.

We believe that a small long-term dependency dataset with high FREQ will not only provide a useful benchmark for language modeling, but also a more suitable testbed for setups like architectural search and meta-learning. We introduce SimpleBooks-92, a dataset of 92M tokens, 90\% that of WikiText-103, but with a vocabulary size of 98K, one third of that of WikiText-103. It has FREQ of 931.4, 90\% that of 1Billion, with OOV tokens accounting for only 0.11\%, even lower than 1Billion. See Table \ref{tab:all-datasets} for comparison.

We also include a 2M-token version, SimpleBooks-2, that has the vocabulary size one third of that of WikiText-2. Transformer models outperform LSTM models on both small and large versions of SimpleBooks.

\begin{table*}[h]
\centering
\scalebox{0.87}{
\begin{tabular}{l|ccccccc}
\toprule
 & \textbf{Source} & \textbf{Tokens} & \textbf{Vocab} & \textbf{Long-term?} & \textbf{OOV} & \textbf{FREQ} & \textbf{SOTA perplexity} \\
 \midrule
1Billion & News & 829M & 793,471 & No & 0.28\% & 1045.09 & 21.8 \citep{dai2019transformer} \\
WikiText-103 & Wikipedia & 103M & 267,735 & Yes & 0.4\% & 385.56 & 16.4 \citep{krause2019dynamic}\\
WikiText-2 & Wikipedia & 2M & 33,278 & Yes & 2.6\% & 62.76 & 39.14 \citep{gong2018frage} \\
PTB & News & 0.9M & 10,000 & No & 4.8\% & 88.75 & 46.54 \citep{gong2018frage} \\ \hline
SimpleBooks-92 & Books & 91.5M & 98,304 & Yes & 0.11\% & 931.4 & 8.921 \\
SimpleBooks-2 & Books & 2.2M & 11,492 & Yes & 0.47\% & 195.43 & 16.407 \\
\bottomrule
\end{tabular}
}
\caption {Comparison between SimpleBooks and popular datasets for word-level LMs} \label{tab:all-datasets} 
\end{table*}

\section{SimpleBooks dataset}
To create this dataset, we downloaded all available books from Gutenberg US (\url{www.gutenberg.org}). After discarding mal-formatted books and books of poems, plays, manuals\footnote{Knitting was apparently a hit in the early 20th century}, recipes, and the literary nonsense, we obtained 39,432 books. We removed meta-data, tables of contents, illustrations. We tokenized the books by simply separating the words by space. Let $L$ be the number of tokens in a book and $V$ be its vocabulary size. Our goal is to choose a subset of those books such that when combining those books together, we have a body of text of approximately 100M tokens but with a vocabulary size of less than 100K.

\begin{figure}[t]
  \centering
  \includegraphics[scale=0.37]{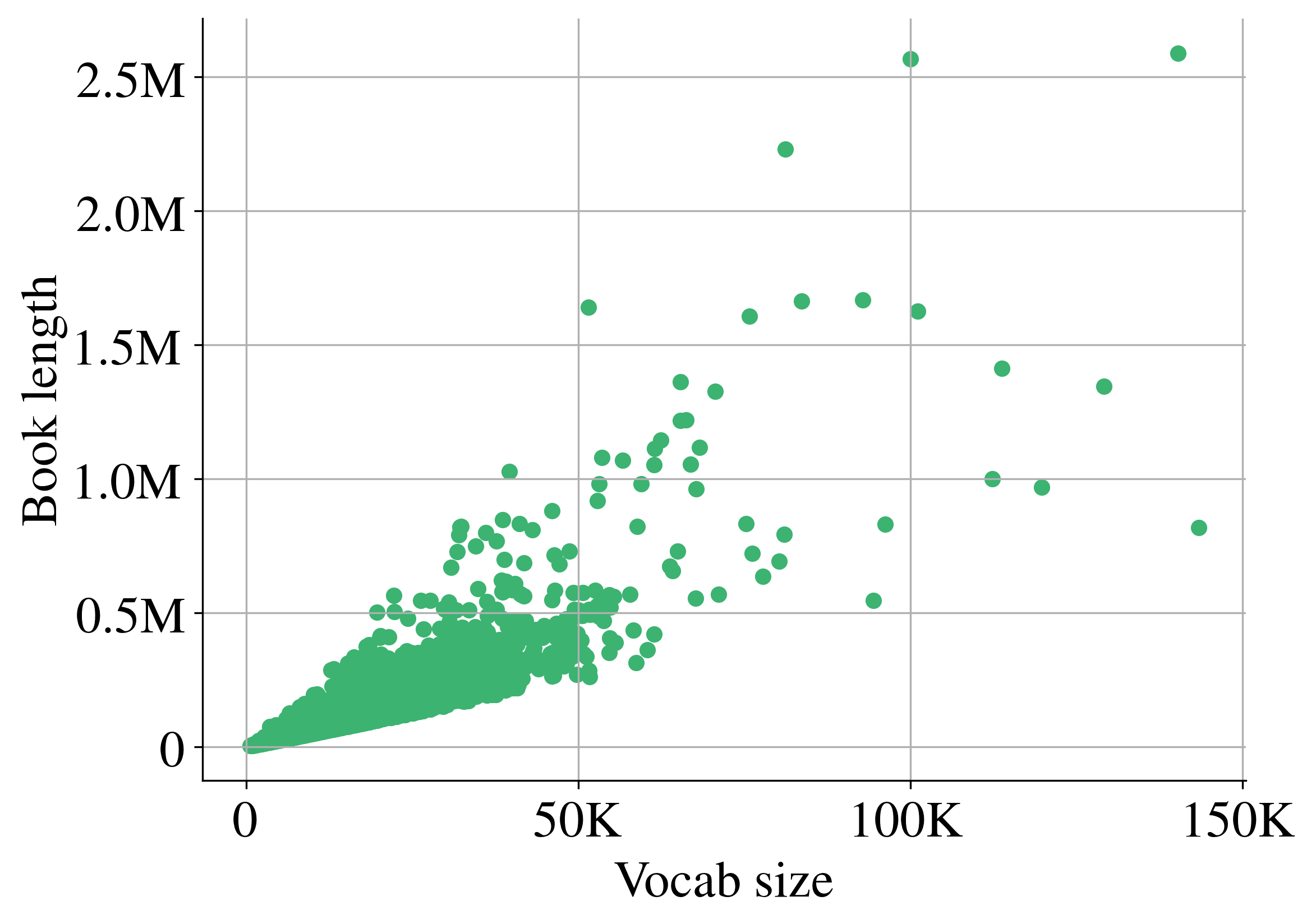}
  \vspace{-3mm}
  \caption{Book length/vocab ratio of Gutenberg books}
  \label{fig:all-books-ratio}
\end{figure}

To do so, we originally chose books with high $\frac{L}{V}$ ratio (which is FREQ). However, this biases towards long books because they tend to have higher FREQ. So we chose books with high $\frac{L}{V^2}$ instead. See Figure \ref{fig:all-books-ratio} for the distribution of the $\frac{L}{V}$ ratio. 

We picked all books with the ratio $\frac{L}{V^2}$ of at least 0.0012. Most of them are children's books, which makes sense since children's books tend to use simpler English. We then went over each book from the largest to the smallest, either adding it to the to-use list or discard it if it has at least 50\% 8-gram token overlap with the books that are already in the to-use list. We ended up with 1,573 books. Scripts used to create this dataset are from the \textit{lazynlp} library \citep{chiphuyen2019}.

Of these 1,573 books, 5 books are used for the validation set and 5 books for the test set. We tokenized each book using SpaCy \citep{spacy2} and separating numbers like ``300,000" and ``1.93" to ``300 @,@ 000" and ``1 @.@ 93". Otherwise, all original case and punctuations are preserved. SimpleBooks-92 contains 92M tokens for train set, and 200k tokens for each validation and test sets. SimpleBooks-2 has the same validation and test sets as SimpleBooks-92, but with only 2M tokens for the train set. %See Table \ref{tab:sb-stats} for SimpleBooks statistics.

We also include the raw version of unprocessed text for character-level LMs.

% \begin{table}[h]
% \centering
% \scalebox{0.5}{
% \begin{tabular}{l|lllllllll}
% \toprule
%  & Train & Vocab & OOV Train & Valid & OOV Valid & Test & OOV Test \\ \hline
% SB-2 & 2,245,899 & 11,492 & 0.47\% & 214,028 & 0.8\% & 209,991 & 0.69\% \\
% SB-92 & 91,560,770 & 98,304 & 0.11\% & 214,028 & 0.07\% & 209,991 & 0.05\% \\
% \bottomrule
% \end{tabular}
% }
% \caption {SimpleBooks statistics} \label{tab:sb-stats} 
% \end{table}

\section{Experiments}
\subsection{Language modeling}
We trained word-level LMs on SimpleBooks-2 and SimpleBooks-92 using both AWD-LSTM\footnote{We used the implementation at \url{https://github.com/salesforce/awd-lstm-lm}} \citep{merity2017regularizing} and Transformer-XL\footnote{We used the implementation at \url{https://github.com/kimiyoung/transformer-xl}} \citep{dai2019transformer}. Note that AWD-LSTM is a highly regularized version of LSTM while the only regularization Transformer-XL uses is dropout.

\begin{table}[h]
\centering
\scalebox{0.9}{
\begin{tabular}{c|cc|cc}
\toprule
\multicolumn{1}{c}{} & \multicolumn{2}{c}{SB-2} & \multicolumn{2}{c}{SB-92} \\
\midrule
\textbf{Model} & \textbf{Valid} & \textbf{Test}  & \textbf{Valid} & \textbf{Test} \\
\midrule
AWD-LSTM & 17.16 & 16.78 & 21.45 & 20.64 \\
Transformer-XL & 17.27 & 16.41 & 9.3 & 8.92 \\
\bottomrule
\end{tabular}
}
\caption {Validation and test perplexities on SB-2 and SB-92 of our best AWD-LSTM and Transformer-XL models.} \label{tab:sb-sota} 
\end{table}

\subsubsection{LSTM vs Transformer on SimpleBooks-2}
We evaluated whether on a small dataset with high FREQ, a vanilla implementation of Transformer models can outperform RNNs, consistent with the results on much larger datasets. We used Milano\footnote{\url{https://github.com/NVIDIA/Milano}} to search through 500 sets of hyperparameters on the first 30 epochs of SimpleBooks-2 for both AWD-LSTM and Transformer-XL. We then trained each architecture on the best set of hyperparameters until convergence. For the set of hyperparameters that we used, see Appendix \ref{hp}.

We found that Transformer-XL indeed outperformed AWD-LSTM on SimpleBooks-2 (See Table \ref{tab:sb-sota}), while also requiring less parameters ($19.7$M against $29.2$M) and fewer epochs to converge ($80$ against $200$).
% \begin{table}[h]
% \centering
% \scalebox{0.7}{
% \begin{tabular}{l|lllll}
% \toprule
%  & Params & Epochs & Hrs & Valid & Test \\ \hline
% AWD-LSTM & 29.2M & 200 & 3 & 17.16 & 16.78 \\
% Transformer-XL & 19.7M & 80 & 1 & 17.266 & 16.407 \\
% \bottomrule
% \end{tabular}
% }
% \caption {The best performance on SimpleBooks-2} \label{tab:sb2-best} 
% \end{table}

\subsubsection{WikiText-103 vs SimpleBooks-92}
It is not surprising that on SimpleBooks-92, both AWD-LSTM and Transformer-XL converge faster and require less parameters compared to on WikiText-103. With identical settings that lead to near-SOTA validation perplexity on both datasets, SimpleBooks-92 can reduce 45.3\% parameters for Transformer-XL and 39.7\% for AWD-LSTM (See Table \ref{tab:training-stats}). Note that both models tie the embedding and softmax layers.

% \begin{table}[h]
% \centering
% \scalebox{0.56}{
% \begin{tabular}{lllllll}
% \toprule
% Model & \textbf{AWD-LSTM} & & & & & \\ \hline
%  & Emb params & Params & Val PPL & Steps & Epoch & Hrs \\ \hline
% WK-103 & - \footnote{Tied weights so the embedding layer is tied to the softmax layer} & 225M & & & &  \\
% SB-92 & - & 94M & 9.594 & 19 & 49 \\\hline
% Model & \textbf{Transformer-XL} & & & & \\\hline
%  & Emb params & Params & Val PPL & Steps & Epoch & Hrs \\ \hline
% WK-103 & 110M & 151M & 26.775 & 80K & 7 & 26 \\
% SB-92 & 40M & 81M & 11.585 & 80K & 8 & 20 \\\hline
% \bottomrule
% \end{tabular}
% }
% \caption {Training statistics on 1 GV100 GPU} \label{tab:training-stats} 
% \end{table}
\begin{table}[h]
\centering
\scalebox{0.9}{
\begin{tabular}{c|cc|cc}

\toprule
\multicolumn{1}{c}{} & \multicolumn{2}{c}{AWD-LSTM} & \multicolumn{2}{c}{Transformer-XL} \\
\midrule
\textbf{Dataset} & \textbf{\# emb} & \textbf{\# params}  & \textbf{\# emb} & \textbf{\# params} \\
\midrule
WK-103 & 128.5M & 205.3M & 137M & 192M  \\
SB-92 & 47.2M & 123.8M & 50.4M & 105M \\
\midrule
WT-2 & 19.2M & 41.8M & 10.6M & 26.6M  \\
SB-2 & 6.6M & 29.2M & 3.7M & 19.7M \\
\bottomrule
\end{tabular}
}
\caption {Number of parameters in models that achieve near-SOTA results on WikiText and SimpleBooks. A large portion of the parameters in WikiText models is concentrated in the embedding layers.} \label{tab:training-stats} 
\end{table}
\subsection{Transfer learning from SimpleBooks to WikiText}
One interesting note is that even though SimpleBooks-92 has the vocabulary size of only 36.7\% that of WikiText-103, it covers 92\%, or 93\% uncased, of all tokens in a slightly different tokenized version of WikiText-103\footnote{In the public copy of WikiText-103, negation contraction such as ``don't" is tokenized as ``don 't". We re-tokenized it as ``do n't" to be consistent with SimpleBooks-92}. This raises a research question: can what we learn from text of simplified English (SimpleBooks-92) be transferred to tasks using normal English (WikiText-103)?

We experimented with training word-embeddings using word2vec skip-gram algorithm \citep{mikolov2013distributed}. We first trained a skip-gram model on SimpleBooks-92 for 100k steps. We then ran two experiments on WikiText-103, each for 200k steps:
\begin{enumerate}
    \item Train a skip-gram model on WikiText-103 from scratch.
    \item For the words in WikiText-103 that are also in SimpleBooks-92, initialize the corresponding rows with the learned embedding from SimpleBooks-92. For all the other rows, uniform randomly initialize them within the (\textit{min}, \textit{max}) range, with \textit{min} being the smallest value in the learned SimpleBooks-92 embedding, and \textit{max} being the largest.
\end{enumerate}

\begin{figure}[t]
  \centering
  \includegraphics[scale=0.23]{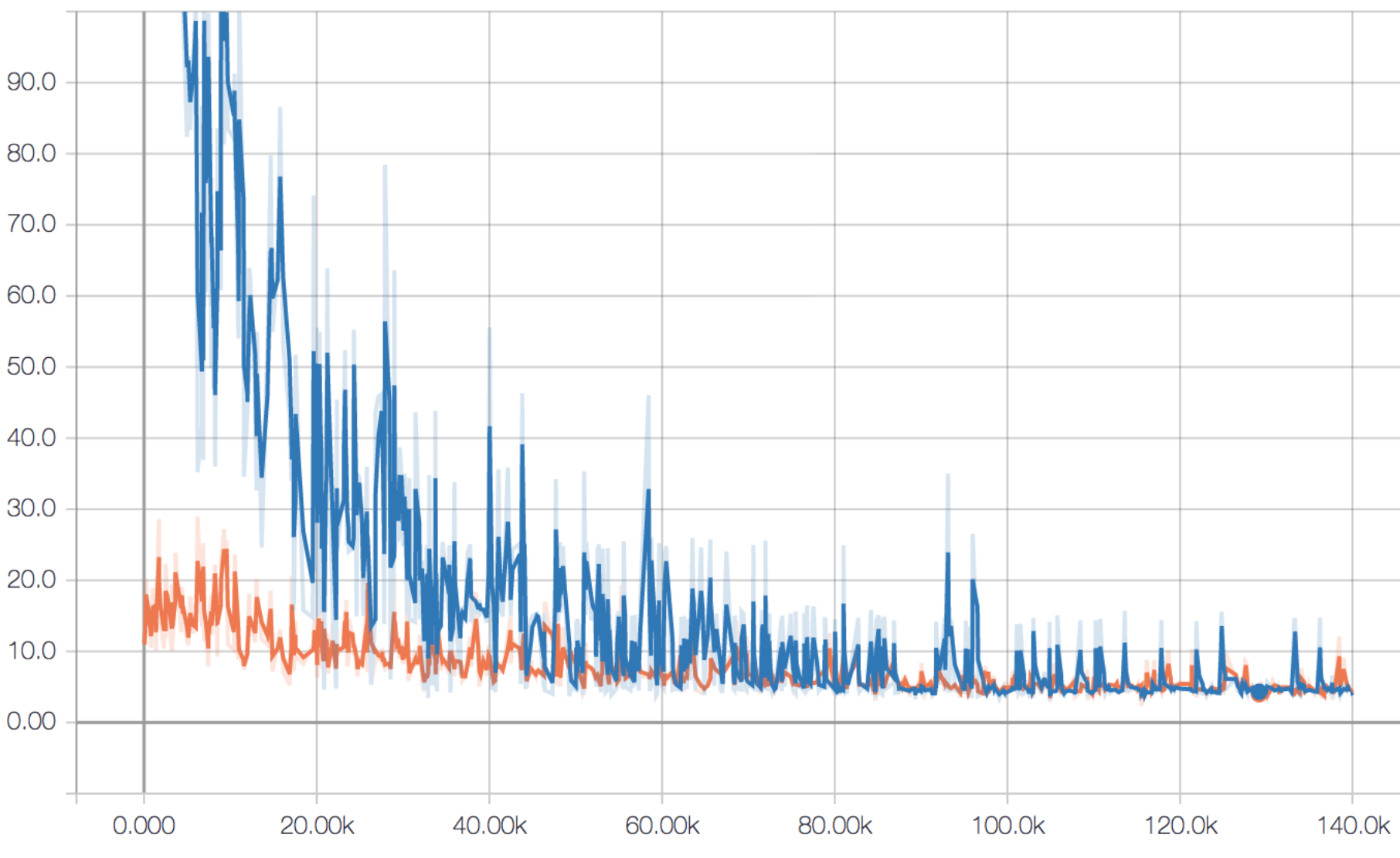}
  \vspace{-3mm}
  \caption{Initializing the embedding matrix from scratch (blue) vs initializing with the embedding matrix trained on SimpleBooks-92 (orange).}
  \label{fig:transfer}
\end{figure}

We found that the second experiment, while the model is able to learn much better, the final losses for both models are comparable (See Figure \ref{fig:transfer}).

\section{Conclusion}
We introduced SimpleBooks-2 and SimpleBooks-92, a 2-million token and a 92-million token dataset, with unique property: they have a  much smaller word-level vocabulary than the current datasets of the same size. This property makes it faster and easier to train word-level LMs on these datasets to convergence, which makes them ideal benchmarks and testbeds for the task of language modeling. While Transformer models usually outperform RNNs on large datasets but underperform RNNs on small datasets, in our experiments, Transformer-XL outperformed AWD-LSTM on both SimpleBooks-2 and SimpleBooks-92.

We also experimented with transfer learning from simple English to normal English with the task of training word embedding and saw some potential. In the future, we would like to experiment with whether it would save time to train a language model on simple English first and use the learned weights to train a language model on normal English.
\section{Acknowledgment}
I'd like to thank my wonderful colleagues Boris Ginsburg, Oleksii Kuchaiev, and Oleksii Hrinchuk for helping me with this project! 
\pagebreak
\bibliography{simplebooks}
\bibliographystyle{unsrtnat}

\appendix
\pagebreak
\appendixpage
\section{Hyperparameters used for training on SimpleBooks-2}\label{hp}
\subsection{AWD-LSTM}
\begin{verbatim}
- alpha: 2.0
- batch_size: 64
- beta: 1.0
- bptt: 48
- clip: 0.9431390850687401
- dropout: 0.09351714464370996
- dropoute: 0.15413135362263264
- dropouth: 0.2379440016364301
- dropouti: 0.782495906512577
- emsize: 576
- lr: 18.0
- nhid: 1152
- nlayers: 3
- nonmono: 5
- optimizer: sgd
- seed: 1882
- tied: True
- wdecay: 1.2e-06
- wdrop: 0.2983586710139643
\end{verbatim}
\subsection{Transformer-XL}
\begin{verbatim}
- n_layer : 12
- n_head : 10
- d_head : 40
- d_embed : 320
- d_model : 320
- d_inner : 1280
- dropout : 0.35
- dropatt : 0.35
- init_range : 0.1
- emb_init_range : 0.01
- init_std : 0.02
- proj_init_std : 0.01
- optim : adam
- lr : 0.00025
- decay_rate : 0.5
- lr_min : 0.0
- clip : 0.25
- clip_nonemb : False
- max_step : 20000
- batch_size : 32
- tgt_len : 150
- eval_tgt_len : 150
- mem_len : 150
- not_tied : False
- seed : 1111
- pre_lnorm : False
- attn_type : 0
- clamp_len : -1
\end{verbatim}
\end{document}